\documentclass[sigconf, nonacm, balance=false]{acmart}

\AtBeginDocument{%
  }

\setcopyright{acmlicensed}
\copyrightyear{2026}
\acmYear{2026}
\acmDOI{XXXXXXX.XXXXXXX}
\acmConference[PILA '26]{Personal Intelligence in the Agentic AI Era Workshop at KDD '26.}{August 9, 2026}{Jeju, Republic of Korea}
\acmISBN{978-1-4503-XXXX-X/2018/06}

\usepackage{enumitem}
\usepackage{multirow}
\usepackage{xurl}
\usepackage{microtype}

\makeatletter
\renewcommand*{\@fnsymbol}[1]{\ensuremath{\ifcase#1\or \hbox{}\or *\or \dagger\or \ddagger\or
   \mathsection\or \mathparagraph\or \|\or **\or \dagger\dagger
   \or \ddagger\ddagger \else\@ctrerr\fi}}
\makeatother
\begin{document}





\title{Situation Graph Prediction for User Perspective Modeling}
\titlenote{Accepted to PILA 2026: Workshop on Personal Intelligence in the Agentic AI Era, at KDD 2026.}
\pagestyle{plain}




\author{Jisung Shin}
\authornote{These authors contributed equally.}
\affiliation{
  \institution{Flybits Labs, Creative AI Hub}
  \institution{University of Toronto}
  \country{}
}
\email{chris.shin@flybits.com}

\author{Daniel Platnick}
\authornotemark[2]
\affiliation{
  \institution{Flybits Labs, Creative AI Hub}
  \institution{Toronto Metropolitan University}
  \country{}
}
\email{daniel.platnick@flybits.com}

\author{Marjan Alirezaie}
\affiliation{%
 \institution{Flybits Labs, Creative AI Hub}
 \institution{Toronto Metropolitan University}
 \country{}
}
\email{marjan.alirezaie@flybits.com}

\author{Hossein Rahnama}
\affiliation{%
   \institution{Flybits Labs, Creative AI Hub}
   \institution{Toronto Metropolitan University}
   \country{}
}
\affiliation{%
  \institution{MIT Media Lab}
  \country{}
}
\email{rahnama@mit.edu}


\renewcommand{\shortauthors}{Shin et al.}


\begin{abstract}
Perspective-aware AI requires modeling evolving internal states---goals, emotions, contexts---not merely preferences.
Progress is limited by a data bottleneck: digital footprints are privacy-sensitive and perspective states are rarely labeled.
We propose \emph{Situation Graph Prediction} (SGP), a task that frames user perspective modeling as an inverse inference problem: reconstructing structured, ontology-aligned representations of perspective from observable multimodal artifacts, suitable as long-horizon memory for personal agents.
To enable grounding without real labels, we use a structure-first synthetic generation strategy that aligns latent labels and observable traces by design.
As a pilot, we construct a dataset and run a diagnostic study using retrieval-augmented in-context learning as a proxy for supervision.
In our diagnostic study across three frontier foundation models (GPT-4o, Gemini 2.5 Flash, Claude Sonnet 4), we observe a consistent positive gap between surface-level extraction and latent perspective inference---indicating that latent-state inference is consistently harder than surface extraction under our controlled setting.
Results suggest SGP is non-trivial and provide evidence for the structure-first data synthesis strategy.
For reproducibility and transparency, our code, data, and supplementary resources are available here: \url{https://github.com/flybits/sg-prediction}.
\end{abstract}




\maketitle

\section{Introduction}
\label{sec:introduction}
Recent advances in foundation models yield impressive general-purpose reasoning and generation capabilities, fueling the rise of agentic AI systems deployed as personal collaborators. However, these systems remain fundamentally impersonal: they reason about the world, but not from the standpoint of a specific individual or organization. As a result, current personal AI agents struggle to act as trustworthy collaborators in domains where understanding evolving goals, values, emotions, and context is essential.

This limitation motivates the emerging paradigm of \emph{Perspective-Aware AI} (PAi) \cite{Alirezaie-pai1-2024}, which shifts the focus from generic personalization toward modeling how an entity experiences and interprets situations over time. 
Rather than modeling users as static preference vectors or through isolated interactions, PAi represents identity as a longitudinal, structured trajectory shaped by lived experience. Such structured representations can serve as a queryable, long-horizon memory substrate for personal agents, supporting user-centric applications such as adaptive education, health support, explainable decision-making, and bias auditing.

Progress in PAi is constrained by a fundamental data bottleneck: longitudinal digital footprints are siloed and privacy-sensitive, and the latent perspective variables underlying behavior (goals, affect, interpretation) are rarely labeled.
Without grounded supervision, training and evaluating an agent's ability to learn a mapping from observable evidence to latent perspective remain intractable.
Consider a personal AI agent that recognizes---from voice tremors, sparse replies, and avoidance patterns---that a user is spiraling toward crisis while behavioral metrics report only ``decreased engagement.''
The motivation of our work is to close this gap while highlighting the ethical obligations that such inference creates.

We advance PAi through three contributions:
\begin{enumerate}[noitemsep, topsep=0pt, leftmargin=*]
    \item We formalize \emph{Situation Graph Prediction} (SGP): a structured inverse inference task mapping observable user data artifacts to ontology-aligned perspective representations suitable as structured memory units for personal agents.
    \item We propose structure-first synthetic generation as a privacy-preserving approach to labeled perspective data.
    \item Through a pilot study spanning three frontier foundation models from different model families (OpenAI GPT-4o, Google Gemini 2.5 Flash, and Anthropic Claude Sonnet 4), we provide evidence that SGP is non-trivial and that latent inference is consistently harder than surface extraction across the three models.
\end{enumerate}

We release our dataset, code, and supplementary resources to enable reproducibility and 
support future research towards larger-scale settings.

\section{Related Work}
\label{sec:related_work}

Our work bridges personalization, long-term identity modeling, and structured multimodal inference.

\paragraph{\textbf{From Personalization to Perspective-Aware AI}}

Classical user modeling represents users as static profiles or preference vectors, enabling surface-level adaptation but not reasoning about evolving internal states \cite{kobsa2001generic}.
Persona-conditioned dialogue improves speaker consistency but treats identity as lightweight text \cite{li-etal-2016-persona, zhang-etal-2018-personalizing}, while personalized alignment emphasizes behavioral adaptation via preferences and controllability \cite{guan-etal-2025-survey}. 
These approaches primarily optimize output tailoring and rarely model how internal states evolve across situations. 
In contrast, PAi targets structured longitudinal representations for personal agents, framing identity as a trajectory of situation-level states (context, affect, goals) rather than isolated preferences \cite{Alirezaie-pai1-2024}. Neurosymbolic user mental modeling similarly constructs ontology-aligned identity graphs from multimodal traces \cite{Rahnama2021Exchangeable,alirezaie-pai2-2024}.

\textbf{Longitudinal Memory and Identity Graphs.}
Long-term memory is central to coherent personal agents. Generative Agents store raw experiences for planning \cite{park2023generativeagentsinteractivesimulacra}, while memory-augmented systems such as MemGPT \cite{packer2023memgpt} and A-Mem \cite{xu2025amem} extend agent context via retrieval and structured memory organization. These approaches focus on textual recall and do not enforce ontology-aligned, queryable perspective representations.
Prior work on PAi introduces \emph{Chronicles} as temporally coherent identity knowledge graphs \cite{Alirezaie-pai1-2024}, with situation graphs representing distinct situational states.
Identity-grounded generation has been shown to reduce persona drift \cite{platnick2025idragidentityretrievalaugmentedgeneration}, and Chronicle-based systems further demonstrate narrative-grounded personalization \cite{platnick2025perspectiveawareaiextendedreality}.

\textbf{Situation Understanding and Latent Mental States.}
Our formulation relates to narrative and commonsense inference. Narrative event chains capture typical event sequences \cite{chambers-jurafsky-2008-unsupervised}, while ATOMIC and COMET model event-intent-reaction relations \cite{sap2019atomicatlasmachinecommonsense, bosselut-etal-2019-comet}. These resources provide useful priors but do not offer instance-level supervision grounded in a specific user's multimodal footprint.

\textbf{Grounded Dynamic Graph Prediction.}
Scene graph generation extracts object--relation structure from images \cite{platnick-pai3-2024}, but does not recover situational structure from multimodal artifacts. Temporal knowledge graph completion models evolving relational data \cite{Cai_2023}, but assumes access to structured inputs rather than inferring them from raw traces. Structured grounding approaches such as K-BERT and REALM motivate the use of graph-based representations for improving reasoning and factuality \cite{liu2019kbertenablinglanguagerepresentation, guu2020realmretrievalaugmentedlanguagemodel}.

\section{Situation Graph Prediction}
In this section we formalize the task of Situation Graph Prediction (SGP), which aims to recover situation-level structured perspective representations from observable digital traces.

\subsection{Situation Graph Representation}
The structured representation $G_t$ is grounded in the DOLCE Ultralite (DUL) upper ontology \cite{StefanoBorgo-2022-dul} and its application to PAi, which provides a foundational vocabulary for describing situations, participants, events, and their relationships. 
We instantiate a domain-specific schema tailored for perspective-aware identity modeling, treating the ontology as a representation of perspective that is sufficient for task grounding.

\textbf{Representation}.
A Situation Graph $G_t = (V_t, \mathcal{R}_t)$ is equivalently represented as a set of semantic triplets
$\mathcal{T}_t = \{(s, p, o) : s, o \in V_t,\; p \in \mathcal{R}_t\}$,
where $s$ is a subject node, $p$ is a predicate (edge label), and $o$ is an object node. Each node $v \in V_t$ is a tuple $(\kappa, \nu)$ comprising a \emph{kind} $\kappa$ drawn from a fixed type vocabulary and a categorical \emph{name} $\nu$ drawn from a type-specific enumeration.
The complete node and edge taxonomies are provided in the supplementary materials; we summarize the key design principles below.

\textbf{Taxonomy and Structural Constraints}.
The schema defines 11 node kinds and 14 edge types organized into four semantic strata:
\emph{participants}, \emph{spatio-temporal structure}, \emph{contextual atmosphere}, and \emph{psychological state}.
Crucially, the schema distinguishes \emph{surface} from \emph{latent} attributes.
Surface nodes (participants, location types, times, ambience) represent observable facets that can plausibly manifest in artifacts, while latent predicates (\texttt{feels}, \texttt{evokes}, \texttt{has\_valence}, \texttt{conveys\_val}) connect to psychological states (\textsc{Emotion}, \textsc{Valence}) encoding unobservable internal perspective.

Well-formedness is enforced through typed constraints: each triplet satisfies
$(s,p,o)\in\mathcal{T}_t \Rightarrow (\kappa_s,\kappa_o)\in\mathcal{A}(p)$,
where $\mathcal{A}(p)$ specifies valid (subject-kind, object-kind) pairs.
Additional structural constraints enforce bounded graph size and minimal completeness conditions (see supplementary materials).

\subsection{Problem Formulation}

We formalize SGP as a latent variable inference task.
At a given time segment $t$, let $Z_t$ denote the user's true latent cognitive state (encompassing internal goals, affect, and context).
Because $Z_t$ is abstract and unobservable, we define the \emph{Situation Graph} $G_t \in \mathcal{G}$ as a structured, symbolic approximation aligned with our ontology, where $\mathcal{G}$ denotes the space of valid graphs under the schema.

The user's state manifests externally through a set of observable digital artifacts $X_t$ (e.g., logs, audio, images, social media posts).
For computational tractability, we model the generative process by treating the structured approximation $G_t$ as the latent variable governing artifact generation.
Specifically, we assume artifacts are drawn from a conditional distribution $X_t \sim P(\cdot \mid G_t)$.
Under this formulation, $X_t$ constitutes a partial, noisy, and unstructured projection of the user's underlying perspective structure encoded by $G_t$.

\subsection{The Inference Task}
In a real-world setting, we only observe $X_t$. 
The goal of SGP is to invert the generative process to recover the perspective structure $G_t$. 
To enable learning this inversion despite the absence of naturally occurring labels, we rely on a \emph{synthetic} proxy dataset. 
Let $\mathcal{D} = \{(X_i, G_i)\}_{i=1}^N$ denote a proxy dataset of pairs where $G_i$ is a ground-truth Situation Graph and $X_i$ is the corresponding set of multimodal digital artifacts. We define the inference objective as:
\[
\hat{G}_t = \operatorname*{argmax}_{G \in \mathcal{G}} P(G \mid X_t; \Theta),
\]
where $\Theta$ represents the inference configuration. 

While this objective can be approached using different learning paradigms (e.g., unsupervised or semi-supervised learning), in this work we focus on two primary regimes:

1) \textbf{Supervised Learning:} $\Theta = \theta$ represents trainable model parameters, and the inference model is instantiated as a function $f_\theta$ such that predictions are given by $\hat{G}_i = f_\theta(X_i)$. The parameters are optimized by minimizing a task-specific loss $\mathcal{L}(f_\theta(X_i), G_i)$ over the dataset $\mathcal{D}$.

2) \textbf{In-Context Inference:} $\Theta = \{\theta_{\text{frozen}}, \mathcal{S}_t\}$, where $\theta_{\text{frozen}}$ denotes a fixed pre-trained foundation model and $\mathcal{S}_t \subset \mathcal{D}$ is a support set dynamically retrieved by relevance (e.g., similarity) to the input $X_t$, to be provided as in-context demonstrations.

This formulation enables principled evaluation of SGP using frozen models. \textit{Note that in our experiments, we focus exclusively on the in-context inference regime to provide evidence for the task design on the pilot dataset, while leaving fully supervised training to future large-scale settings.}

\subsubsection{Bridging the Supervision Gap.}
A critical challenge in PAi is that real-world digital footprints lack explicit ground truth labels $G_t$; users do not annotate their lives with knowledge graphs. Consequently, standard supervision is impossible on real data.

To address this, we adopt a synthetic supervision approach.
We use a synthetic generator to sample aligned pairs $(G_t, X_t)$ from the ontology's prior distribution, where $G_t$ specifies a valid situation structure and $X_t$ consists of artifacts conditionally rendered to be consistent with that structure. 
We detail this approach in Section \ref{sec:struct-first-synth-data}.

\subsubsection{SGP Task Decomposition}
We decompose SGP into two subtasks to isolate distinct modeling challenges:

\textbf{Task 1:} \textit{Static Situation Graph Prediction.}
The static task evaluates multimodal grounding in isolation, requiring the model to infer $G_t$ from a single time segment's artifacts without access to historical context:
$\hat{G}_t = f_{\text{static}}(X_t)$.
This setting measures a model's ability to integrate heterogeneous signals (text, images, audio, logs) into a unified structured representation consistent with the ontology.

\textbf{Task 2:} \textit{Temporal Situation Graph Prediction.}
The temporal task evaluates longitudinal reasoning, including causal consistency and narrative continuity. To disentangle reasoning capability from error accumulation, we consider two evaluation modes:

1) \textbf{Single-Step Transition (Oracle History).}  
The model estimates $\hat{G}_t = \operatorname*{argmax}_{G} P(G \mid X_t, H_t; \Theta)$, leveraging the ground-truth history $H_t = (G_{t-k}, \ldots, G_{t-1})$ to constrain the latent state transition.
This mode isolates the model's ability to infer state transitions under ideal historical context.

2) \textbf{Multi-Step Trajectory (Autoregressive).}  
The model predicts $\hat{G}_t = \arg\max_G P(G \mid X_t, \hat{H}_t; \Theta)$, where 
$\hat{H}_t = (\hat{G}_{t-k}, \dots, \hat{G}_{t-1})$ are the model's prior predictions (with context window size $k$).
This setting evaluates robustness to error accumulation and the ability to maintain coherent perspective over time.

Note that we evaluate Task 1 exclusively, as reliable single-timestep grounding is a prerequisite before temporal error accumulation can be meaningfully studied.

\subsubsection{Output Representation and Evaluation}
Because $G_t$ is isomorphic to a set of semantic triplets $\mathcal{T}_t$, the model output is defined as a predicted set $\hat{\mathcal{T}}_t$ drawn from the space of ontology-compliant triplets. 
For in-context inference, we evaluate performance using standard set-based metrics (precision, recall, $F_1$) between the predicted set $\hat{\mathcal{T}}_t$ and ground truth $\mathcal{T}_t$ \cite{powers2011evaluation}.


\section{Structure-First Synthetic Data Generation}
\label{sec:struct-first-synth-data}
To train and evaluate SGP models, we require paired examples $(X_t, G_t)$, where $X_t$ denotes multimodal digital artifacts and $G_t$ the corresponding ground-truth Situation Graph.
Because real-world data lacks explicit $G_t$ labels, we adopt a \emph{structure-first} synthetic generation pipeline.
Rather than prompting a language model to freely simulate users, we explicitly invert the generative process.

First, we sample a valid Situation Graph $G_t$ from the predefined ontology, encoding entities, relations, goals, and affect for a hypothetical user in a given context (e.g., \emph{Job Interview}, \emph{Family Conflict}).
Next, we treat artifact generation as a conditional rendering task.
An LLM is prompted to produce observable evidence $X_t$ (e.g., emails, chat logs, calendar entries) that is constrained to be consistent with the structure encoded in $G_t$.
This inversion aligns supervision with evidence by construction: artifacts are grounded in the perspective structure rather than annotated post hoc.
The resulting dataset enables controlled, privacy-preserving study of SGP without relying on real user data.

\section{Pilot Data Construction} 
To test the structure-first pipeline and instantiate the SGP task, we constructed a focused pilot dataset of synthetic situations. The dataset and supporting resources are available at \url{https://github.com/flybits/sg-prediction}, including code to reproduce our experiments. 
Data were generated using the pipeline described in Section~\ref{sec:struct-first-synth-data} and reviewed by a human expert to ensure consistency and quality.
The dataset contains $N=75$ situation instances across multiple domains (e.g., professional, personal, health-related) and modalities (primarily text, with a subset including images and audio), comprising 225 unique synthetic artifacts. Each instance is a paired $(X_t, G_t)$ sample, where $G_t$ is a ground-truth Situation Graph and $X_t$ the corresponding set of artifacts.

All data are centered on a fictional persona, \emph{Elise Navarro}, a 28-year-old Filipino professional living in Toronto and working as a Senior Marketing Analyst. The dataset spans 75 temporally ordered events across 2021--2025 ($\sim$60 months), covering four domains: \textit{professional developments}, \textit{personal and lifestyle changes}, \textit{health and physical milestones}, and \textit{social and relational experiences}. Events were sampled via the structured generation pipeline and manually curated into a coherent longitudinal timeline to support temporal consistency. 
We use a single persona to ensure longitudinal coherence; multi-persona scaling follows the same methodology without changing the task formulation.

\begin{table*}[t]
\centering
\caption{SGP results across three foundation models (Zero-Shot vs. RA-ICL). Each row is a 5-fold CV mean averaged over 3 independent runs ($\pm$ mean within-run across-fold std). Soft $F_1$ uses cosine similarity on \textit{text-embedding-3-large} with threshold 0.8; the entropy-normalized gap adjusts for vocabulary diversity (7 latent vs.\ 101 surface object values observed in the pilot).}
\label{tab:results}
\vspace{-6pt}
\begin{tabular}{l l c c c c}
\toprule
\textbf{Model} & \textbf{Mode} & \textbf{Strict $F_1$ $\uparrow$} & \textbf{Soft $F_1$ $\uparrow$} & \textbf{PVR $\downarrow$} & \textbf{Gap$^{*}$ $\uparrow$} \\
\midrule
\multirow{2}{*}{GPT-4o}
  & Zero-Shot & 0.016 ($\pm$0.015) & 0.145 ($\pm$0.072) & 0.061 ($\pm$0.011) & +0.139 ($\pm$0.134) \\
  & RA-ICL & 0.163 ($\pm$0.081) & 0.424 ($\pm$0.097) & 0.065 ($\pm$0.024) & \textbf{+0.528} ($\pm$0.359) \\
\midrule
\multirow{2}{*}{Gemini 2.5 Flash}
  & Zero-Shot & 0.020 ($\pm$0.014) & 0.174 ($\pm$0.046) & 0.000 ($\pm$0.000) & +0.210 ($\pm$0.207) \\
  & RA-ICL & 0.221 ($\pm$0.110) & 0.486 ($\pm$0.137) & 0.002 ($\pm$0.003) & \textbf{+0.668} ($\pm$0.303) \\
\midrule
\multirow{2}{*}{Claude Sonnet 4}
  & Zero-Shot & 0.028 ($\pm$0.019) & 0.291 ($\pm$0.111) & 0.000 ($\pm$0.000) & +0.396 ($\pm$0.251) \\
  & RA-ICL & \textbf{0.323} ($\pm$0.124) & \textbf{0.623} ($\pm$0.118) & 0.072 ($\pm$0.004) & \textbf{+0.704} ($\pm$0.385) \\
\bottomrule
\end{tabular}

\parbox{0.95\linewidth}{\footnotesize \vspace{4pt}
$^{*}$Gap = entropy-normalized $F^{*}_1(\text{surface})-F^{*}_1(\text{latent})$.
Positive values indicate latent inference is harder than surface extraction.
Across all three models, RA-ICL substantially improves $F_1$, and the entropy-normalized latent--surface gap is positive in both conditions and grows under RA-ICL---suggesting retrieval aids surface extraction more than latent inference \emph{regardless of model family}.
Claude Sonnet 4 attains the highest $F_1$ and the largest gap; Gemini 2.5 Flash exhibits the lowest predicate-violation rate (PVR).
Full per-model P/R breakdown is provided in the supplementary materials.}
\end{table*}

While intentionally small-scale, this dataset suffices to explore our structure-first approach and demonstrate that SGP poses non-trivial structural and inferential challenges for the evaluated models.

\section{Diagnostic Study}
To assess the difficulty of SGP, we conduct a diagnostic study with an LLM-based pipeline. We evaluate only \textbf{Task 1 (Static SGP)}, focusing on grounding multimodal artifacts into a single-step structured graph, and defer longitudinal modeling (Task 2). Given the pilot size ($N=75$), the goal is not benchmarking, but studying task feasibility and probing the \emph{semantic gap} between surface processing and latent-state inference.
We use an 80/20 stratified split with 5-fold cross-validation (60 retrieval, 15 test), ensuring each instance is evaluated once.

\subsection{Methodology and Experiment Procedure} 
Our pipeline maps raw artifacts $X_t$ to triplets $\hat{\mathcal{T}}_t$ in three stages:

\paragraph{\textbf{1. Modality Decomposition.}} To address the heterogeneity of digital footprints, we first transform raw artifacts into a unified textual representation using modality-specific encoders.
Text is processed directly (e.g., social media posts, chat logs).
Images are converted into descriptive tags and scene summaries via a Vision-Language Model (VLM).
Audio files are processed into exact transcripts and paralinguistic descriptors (e.g., \emph{voice\_tremor}, \emph{loudness}).

\paragraph{\textbf{2. Diagnostic Protocols.}}
We evaluate three frontier foundation models---GPT-4o (OpenAI), Gemini 2.5 Flash (Google), and Claude Sonnet 4 (Anthropic)---under two distinct protocols to isolate the impact of structured supervision:

(1) \textbf{Zero-shot schema alignment.} The model receives decomposed artifacts and full ontology definitions (node/edge taxonomies and constraints) but no in-context examples, measuring its ability to map surface evidence to abstract schema.

(2) \textbf{Retrieval-augmented in-context learning (RA-ICL).} We retrieve the top-$k{=}3$ semantically similar $(X,G)$ pairs using \textit{text-embedding-3-large} and provide them as demonstrations. These serve as proxy supervision, illustrating how artifact patterns (e.g., a curt email and high-tempo audio) map to graph structures (e.g., \textsc{SocialContext} $\Rightarrow$ \textsc{Professional}, \textsc{Emotion}$\Rightarrow$\textsc{Stressed}).
All three models are evaluated under identical prompts, retrieval indices, splits, and seeds.
Per-model handlers and configuration details are provided in the supplementary materials.

\paragraph{\textbf{3. Graph Generation and Evaluation Metrics.}}

The model predicts triplets $\hat{\mathcal{T}}_t$ by fusing multimodal descriptors, schema constraints, and optional demonstrations.
We assess fidelity using four metrics:  
(1) \emph{Predicate Violation Rate (PVR)} for ontological compliance (fraction of invalid predicates);  
(2) \emph{Strict $F_1$} via exact string matching;  
(3) \emph{Soft $F_1$} using embedding similarity (\texttt{text-embedding-3-\allowbreak large}), decomposed into \emph{latent} ($\mathcal{T}^{\text{lat}}$, e.g., \textsc{Emotion}, \textsc{Valence}) and \emph{surface} ($\mathcal{T}^{\text{surf}}$) subsets to separate extraction from state inference; and  
(4) the \emph{latent--surface gap} $\Delta_{\mathrm{LS}}=F_1^{\text{surf}}-F_1^{\text{lat}}$ (positive values indicate surface extraction outperforms latent inference, suggesting latent states are harder to recover).

To control for varying vocabulary sparsity (7 latent vs.\ 101 surface unique object values observed in the pilot dataset), we report an entropy-normalized $F_1^* = F_1 \cdot (H_{\text{cat}} / H_{\text{base}})$, where $H_{\text{cat}}$ is the Shannon entropy of the target category's empirical value distribution in the pilot dataset and $H_{\text{base}}$ is the geometric mean of category entropies serving as the complexity baseline (see supplementary materials).
All results use 5-fold cross-validation with $k{=}3$ retrieved examples per query.

\subsection{Results, Analysis, and Task Feasibility}

Table~\ref{tab:results} summarizes our diagnostic results across the three models.
SGP poses a non-trivial challenge for all three frontier foundation models: while RA-ICL substantially improves overall performance for every model (Soft $F_1$ gains of $+0.28$ for GPT-4o, $+0.31$ for Gemini, $+0.33$ for Claude), all three more readily extract explicit surface elements (e.g., \textsc{Participants}, \textsc{Locations}) than infer latent perspective variables (e.g., \textsc{Emotion}, \textsc{Valence}).
Qualitative examples in the supplementary materials illustrate this distinction, which is quantified by the latent--surface gap.

\textbf{Cross-Model Generalization.}
The central finding---that latent perspective inference is harder than surface extraction---reproduces across all three foundation models.
Under RA-ICL, the entropy-normalized latent--surface gap is positive in every case ($+0.528$ GPT-4o, $+0.668$ Gemini, $+0.704$ Claude), and the same direction holds in the zero-shot condition.
Because the three models come from different organizations with different pre-training corpora and multimodal stacks, the consistency of the gap is unlikely to reflect any single model's idiosyncrasies.

\textbf{Entropy Normalization.}
The latent vocabulary is substantially smaller than the surface vocabulary (8 vs.\ 106 values in the schema; 7 vs.\ 101 observed in the pilot),  which should, in principle, make latent matching easier.
To control for this confound, we apply entropy normalization, scaling each category's $F_1$ by its vocabulary entropy ratio.
For every model, normalization \emph{widens} the latent--surface gap relative to the raw soft-$F_1$ gap, indicating that raw parity is misleading: once we account for vocabulary difficulty, latent inference is substantially harder than surface extraction.

\textbf{Behavioral Profile Differences.}
While the gap generalizes in direction, the three models occupy distinct operating points.
Claude Sonnet 4 attains the highest Strict $F_1$ (0.323) and Soft $F_1$ (0.623) under RA-ICL, alongside the largest normalized gap (+0.704)---suggesting that increased base capability disproportionately benefits surface extraction.
Gemini 2.5 Flash exhibits markedly higher recall than precision (Soft R/P = 0.56/0.44) and produces near-zero predicate violations (PVR = 0.002), an over-generation tendency tempered by strict ontology compliance.
GPT-4o sits between, with balanced precision and recall (Soft R/P = 0.42/0.44) and a moderate violation rate (PVR = 0.065).
These divergent profiles---visible in precision and recall but invisible to $F_1$ alone---underscore that the task probes capabilities beyond simple accuracy.

\section{Discussion, Limitations, and Ethics}
\label{sec:limitations}
SGP provides a concrete task for structured perspective inference, but this work has limitations. 
Our pilot dataset provides evidence for structure-first generation rather than supporting large-scale training, and scaling the pipeline is a key next step. 
While we evaluate SGP via retrieval-augmented in-context inference, the formulation naturally supports unsupervised and semi-supervised paradigms (e.g., treating $G_t$ as latent in a VAE framework).

The structure-first pipeline is limited by the expressivity of the ontology: phenomena not represented in our schema (e.g., culturally specific norms) cannot be captured.
Further, synthetic artifacts lack some of the noise and irregularity of real-world traces---our setting offers a \emph{lower bound} on difficulty, and the gap may widen on noisier real data.
Our study evaluates three frontier closed-weight foundation models, which suggests that the latent--surface gap is not an artifact of a single model; broader evaluation across open-weight models, specialized multimodal architectures, and fine-tuned regimes remains needed to fully disentangle task difficulty from model-specific behavior.

SGP raises ethical considerations due to its focus on inferring internal states from user data.
Our approach prioritizes privacy by relying exclusively on synthetic data and avoiding real user traces. 
Any ontology for human perspective is inherently normative and requires interdisciplinary scrutiny. 
Deployed systems must ensure users retain control over how inferred states inform adaptation.

\section{Conclusion}

We introduced \emph{Situation Graph Prediction} as a task for recovering structured, ontology-aligned perspective representations from multimodal digital artifacts. To enable research under privacy constraints, we proposed a structure-first synthetic generation methodology and instantiated it with a pilot dataset and diagnostic evaluation across three frontier foundation models (GPT-4o, Gemini 2.5 Flash, Claude Sonnet 4).
All three models extract surface-level structure more readily than they recover latent states after controlling for vocabulary diversity. The entropy-normalized latent--surface gap is positive across models and grows under RA-ICL ($+0.53$, $+0.67$, $+0.70$), suggesting the gap is not an artifact of a single model family.
Our work provides a concrete task framework for advancing structured, transparent, and perspective-aware personal agents.

\begin{acks}
The authors wish to express gratitude to the teams at Flybits,
Toronto Metropolitan University, The Creative School, and MIT
Media Lab for their valuable support.
\end{acks}

\bibliographystyle{ACM-Reference-Format}
\bibliography{refs}

@inproceedings{Rahnama2021Exchangeable,
  author    = {Hossein Rahnama and Marjan Alirezaie and Alex Pentland},
  title     = {A Neural-Symbolic Approach for User Mental Modeling: A Step Towards Building Exchangeable Identities},
  booktitle = {AAAI 2021 Spring Symposium on Combining Machine Learning and Knowledge Engineering (MAKE)},
  year      = {2021}
}

@misc{platnick2025perspectiveawareaiextendedreality,
      title={Perspective-Aware AI in Extended Reality}, 
      author={Daniel Platnick and Matti Gruener and Marjan Alirezaie and Kent Larson and Dava J. Newman and Hossein Rahnama},
      year={2025},
      eprint={2507.11479},
      archivePrefix={arXiv},
      primaryClass={cs.AI},
      url={https://arxiv.org/abs/2507.11479}, 
}

@inproceedings{zhang-etal-2018-personalizing,
    title = "Personalizing Dialogue Agents: {I} have a dog, do you have pets too?",
    author = "Zhang, Saizheng  and
      Dinan, Emily  and
      Urbanek, Jack  and
      Szlam, Arthur  and
      Kiela, Douwe  and
      Weston, Jason",
    booktitle = "Proceedings of the 56th Annual Meeting of the Association for Computational Linguistics (Volume 1: Long Papers)",
    month = jul,
    year = "2018",
    address = "Melbourne, Australia",
    publisher = "Association for Computational Linguistics",
    doi = "10.18653/v1/P18-1205",
    pages = "2204--2213",
}

@inproceedings{guan-etal-2025-survey,
    title = "A Survey on Personalized {A}lignment{---}{T}he Missing Piece for Large Language Models in Real-World Applications",
    author = "Guan, Jian  and
      Wu, Junfei  and
      Li, Jia-Nan  and
      Cheng, Chuanqi  and
      Wu, Wei",
    editor = "Che, Wanxiang  and
      Nabende, Joyce  and
      Shutova, Ekaterina  and
      Pilehvar, Mohammad Taher",
    booktitle = "Findings of the Association for Computational Linguistics: ACL 2025",
    month = jul,
    year = "2025",
    address = "Vienna, Austria",
    publisher = "Association for Computational Linguistics",
    url = "https://aclanthology.org/2025.findings-acl.277/",
    doi = "10.18653/v1/2025.findings-acl.277",
    pages = "5313--5333",
    ISBN = "979-8-89176-256-5"
}

@misc{park2023generativeagentsinteractivesimulacra,
      title={Generative Agents: Interactive Simulacra of Human Behavior}, 
      author={Joon Sung Park and Joseph C. O'Brien and Carrie J. Cai and Meredith Ringel Morris and Percy Liang and Michael S. Bernstein},
      year={2023},
      eprint={2304.03442},
      archivePrefix={arXiv},
      primaryClass={cs.HC},
      url={https://arxiv.org/abs/2304.03442}, 
}

@misc{platnick2025idragidentityretrievalaugmentedgeneration,
      title={ID-RAG: Identity Retrieval-Augmented Generation for Long-Horizon Persona Coherence in Generative Agents}, 
      author={Daniel Platnick and Mohamed E. Bengueddache and Marjan Alirezaie and Dava J. Newman and Alex ''Sandy'' Pentland and Hossein Rahnama},
      year={2025},
      eprint={2509.25299},
      archivePrefix={arXiv},
      primaryClass={cs.AI},
      url={https://arxiv.org/abs/2509.25299}, 
}

@inproceedings{Alirezaie-pai1-2024,
  author = {Alirezaie, Marjan and Rahnama, Hossein and Pentland, Alex},
  title = {Structural Learning in the Design of Perspective-Aware AI Systems Using Knowledge Graphs},
  booktitle = {Digital Human Workshop at AAAI Conference on Artificial Intelligence},
  year = {2024},
}

@inproceedings{li-etal-2016-persona,
    title = "A Persona-Based Neural Conversation Model",
    author = "Li, Jiwei  and
      Galley, Michel  and
      Brockett, Chris  and
      Spithourakis, Georgios  and
      Gao, Jianfeng  and
      Dolan, Bill",
    editor = "Erk, Katrin  and
      Smith, Noah A.",
    booktitle = "Proceedings of the 54th Annual Meeting of the Association for Computational Linguistics (Volume 1: Long Papers)",
    month = aug,
    year = "2016",
    address = "Berlin, Germany",
    publisher = "Association for Computational Linguistics",
    url = "https://aclanthology.org/P16-1094/",
    doi = "10.18653/v1/P16-1094",
    pages = "994--1003"
}

@inproceedings{chambers-jurafsky-2008-unsupervised,
    title = "Unsupervised Learning of Narrative Event Chains",
    author = "Chambers, Nathanael  and
      Jurafsky, Dan",
    editor = "Moore, Johanna D.  and
      Teufel, Simone  and
      Allan, James  and
      Furui, Sadaoki",
    booktitle = "Proceedings of ACL-08: HLT",
    month = jun,
    year = "2008",
    address = "Columbus, Ohio",
    publisher = "Association for Computational Linguistics",
    url = "https://aclanthology.org/P08-1090/",
    pages = "789--797"
}

@misc{sap2019atomicatlasmachinecommonsense,
      title={ATOMIC: An Atlas of Machine Commonsense for If-Then Reasoning}, 
      author={Maarten Sap and Ronan LeBras and Emily Allaway and Chandra Bhagavatula and Nicholas Lourie and Hannah Rashkin and Brendan Roof and Noah A. Smith and Yejin Choi},
      year={2019},
      eprint={1811.00146},
      archivePrefix={arXiv},
      primaryClass={cs.CL},
      url={https://arxiv.org/abs/1811.00146}, 
}

@inproceedings{bosselut-etal-2019-comet,
    title = "{COMET}: Commonsense Transformers for Automatic Knowledge Graph Construction",
    author = "Bosselut, Antoine  and
      Rashkin, Hannah  and
      Sap, Maarten  and
      Malaviya, Chaitanya  and
      Celikyilmaz, Asli  and
      Choi, Yejin",
    editor = "Korhonen, Anna  and
      Traum, David  and
      M{\`a}rquez, Llu{\'i}s",
    booktitle = "Proceedings of the 57th Annual Meeting of the Association for Computational Linguistics",
    month = jul,
    year = "2019",
    address = "Florence, Italy",
    publisher = "Association for Computational Linguistics",
    url = "https://aclanthology.org/P19-1470/",
    doi = "10.18653/v1/P19-1470",
    pages = "4762--4779"
}

@article{platnick-pai3-2024,
AUTHOR = {Platnick, Daniel and Alirezaie, Marjan and Rahnama, Hossein},
TITLE = {Enabling Perspective-Aware AI with Contextual Scene Graph Generation},
JOURNAL = {Information},
VOLUME = {15},
YEAR = {2024},
NUMBER = {12},
ARTICLE-NUMBER = {766},
DOI = {10.3390/info15120766}
}

@inproceedings{Cai_2023, series={IJCAI-2023},
   title={Temporal Knowledge Graph Completion: A Survey},
   url={http://dx.doi.org/10.24963/ijcai.2023/734},
   DOI={10.24963/ijcai.2023/734},
   booktitle={Proceedings of the Thirty-Second International Joint Conference on Artificial Intelligence},
   publisher={International Joint Conferences on Artificial Intelligence Organization},
   author={Cai, Borui and Xiang, Yong and Gao, Longxiang and Zhang, He and Li, Yunfeng and Li, Jianxin},
   year={2023},
   month=aug, pages={6545–6553},
   collection={IJCAI-2023} }

@misc{liu2019kbertenablinglanguagerepresentation,
      title={K-BERT: Enabling Language Representation with Knowledge Graph}, 
      author={Weijie Liu and Peng Zhou and Zhe Zhao and Zhiruo Wang and Qi Ju and Haotang Deng and Ping Wang},
      year={2019},
      eprint={1909.07606},
      archivePrefix={arXiv},
      primaryClass={cs.CL},
      url={https://arxiv.org/abs/1909.07606}, 
}

@inproceedings{guu2020realmretrievalaugmentedlanguagemodel,
    author = {Guu, Kelvin and Lee, Kenton and Tung, Zora and Pasupat, Panupong and Chang, Ming-Wei},
    title = {REALM: retrieval-augmented language model pre-training},
    year = {2020},
    publisher = {JMLR.org},
    booktitle = {Proceedings of the 37th International Conference on Machine Learning},
    articleno = {368},
    numpages = {10},
    series = {ICML'20}
}

@article{alirezaie-pai2-2024,
  author = {Marjan Alirezaie and Daniel Platnick and Hossein Rahnama and Alex Pentland},
  title = {Perspective-Aware AI (PAi) for Augmenting Critical Decision Making},
  journal = {TechRxiv},
  year = {2024},
  doi = {10.36227/techrxiv.173602815.51151031/v2},
}

@article{StefanoBorgo-2022-dul,
author = {Stefano Borgo and Roberta Ferrario and Aldo Gangemi and Nicola Guarino and Claudio Masolo and Daniele Porello and Emilio M. Sanfilippo and Laure Vieu},
title ={DOLCE: A descriptive ontology for linguistic and cognitive engineering},

journal = {Applied Ontology},
volume = {17},
number = {1},
pages = {45-69},
year = {2022},
doi = {10.3233/AO-210259},
eprint = { 
    
    
        https://journals.sagepub.com/doi/pdf/10.3233/AO-210259
    

}
}

@article{powers2011evaluation,
  title={Evaluation: From Precision, Recall and F-Measure to ROC, Informedness, Markedness \& Correlation},
  author={Powers, David M. W.},
  journal={Journal of Machine Learning Technologies},
  volume={2},
  number={1},
  pages={37--63},
  year={2011}
}

@article{kobsa2001generic,
author = {Kobsa, Alfred},
year = {2001},
pages = {},
title = {Generic User Modeling Systems},
volume = {11},
isbn = {978-3-540-72078-2},
journal = {User Model. User-Adapted Interact.},
doi = {10.1023/A:1011187500863}
}

@article{packer2023memgpt,
  title   = {MemGPT: Towards LLMs as Operating Systems},
  author  = {Packer, Charles and Wooders, Sarah and Lin, Kevin and Fang, Vivian and Patil, Shishir G. and Stoica, Ion and Gonzalez, Joseph E.},
  journal = {arXiv preprint arXiv:2310.08560},
  year    = {2023}
}

@inproceedings{xu2025amem,
  title     = {A-Mem: Agentic Memory for LLM Agents},
  author    = {Xu, Wujiang and Liang, Zujie and Mei, Kai and Gao, Hang and Tan, Juntao and Zhang, Yongfeng},
  booktitle = {Advances in Neural Information Processing Systems (NeurIPS)},
  year      = {2025}
}


\end{document}